\documentclass[11pt,a4paper]{article}
\usepackage[hyperref]{emnlp2018}
\usepackage{times}
\usepackage{latexsym}
\usepackage{graphicx}
\usepackage{enumitem}
\usepackage{todonotes}
\usepackage{booktabs}
\usepackage{url}
\usepackage{subfig}
\usepackage[export]{adjustbox}
\usepackage[normalem]{ulem} 
\aclfinalcopy 

\title{Distant Supervision from Disparate Sources\\for Low-Resource Part-of-Speech Tagging}

\author{Barbara Plank \and {\v Z}eljko Agi{\' c}\\
	Department of Computer Science\\
    IT University of Copenhagen\\
    Rued Langgaards Vej 7, 2300 Copenhagen S, Denmark\\
    {\tt bplank@itu.dk, zeag@itu.dk}}

\date{}

\begin{document}
\maketitle
\begin{abstract}
We introduce {\sc DsDs}:\ a cross-lingual neural part-of-speech tagger that learns from disparate sources of distant supervision, and realistically scales to hundreds of low-resource languages. The model exploits annotation projection, instance selection, tag dictionaries, morphological lexicons, and distributed representations, all in a uniform framework. The approach is simple, yet surprisingly effective, resulting in a new state of the art {\em without} access to any gold annotated data.

\end{abstract}

\section{Introduction}

Low-resource languages lack manually annotated data to learn even the most basic models such as part-of-speech (POS) taggers. To compensate for the absence of direct supervision, work in cross-lingual learning and distant supervision has discovered creative use for a number of alternative data sources to learn feasible models:
\begin{itemize}[noitemsep,leftmargin=*,label=--,topsep=0.25ex]
\item aligned {\it parallel corpora} to project POS annotations to target languages \cite{yarowsky2001inducing,agic:ea:2015,fang2016learning}, 
\item noisy {\it tag dictionaries} for type-level approximation of full supervision \cite{li-et-al:2012}, 
\item combination of projection and {\it type constraints} \cite{das-petrov:2011,tackstrom:ea:2013},
\item rapid annotation of {\it seed training data} \cite{garrette-baldridge:2013:NAACL-HLT,garrette-mielens-baldridge:2013:ACL2013}.
\end{itemize}

However, 
only one or two compatible sources of distant supervision are typically employed. In reality severely under-resourced languages may require a more pragmatic ``take what you can get'' viewpoint. Our results suggest that combining supervision sources is the way to go about creating viable low-resource taggers. 

We propose a method to strike a balance between model simplicity and the capacity to easily integrate heterogeneous learning signals. Our system is a uniform neural model for POS tagging that learns from {\it disparate sources of distant supervision} ({\sc DsDs}). We use it to combine: i) multi-source annotation projection, ii) instance selection, iii) noisy tag dictionaries, and iv) distributed word and sub-word representations.
We examine how far we can get by exploiting only the wide-coverage resources that are currently readily available for more than 300 languages, which is the breadth of the parallel corpus we employ.

{\sc DsDs} yields a new state of the art by jointly leveraging disparate sources of distant supervision in an experiment with 25 languages. We demonstrate: i) substantial gains in carefully selecting high-quality instances in annotation projection, ii) the usefulness of lexicon features for neural tagging, and iii) the importance of word embeddings initialization for faster convergence. 

\begin{figure}
\includegraphics[width=\columnwidth]{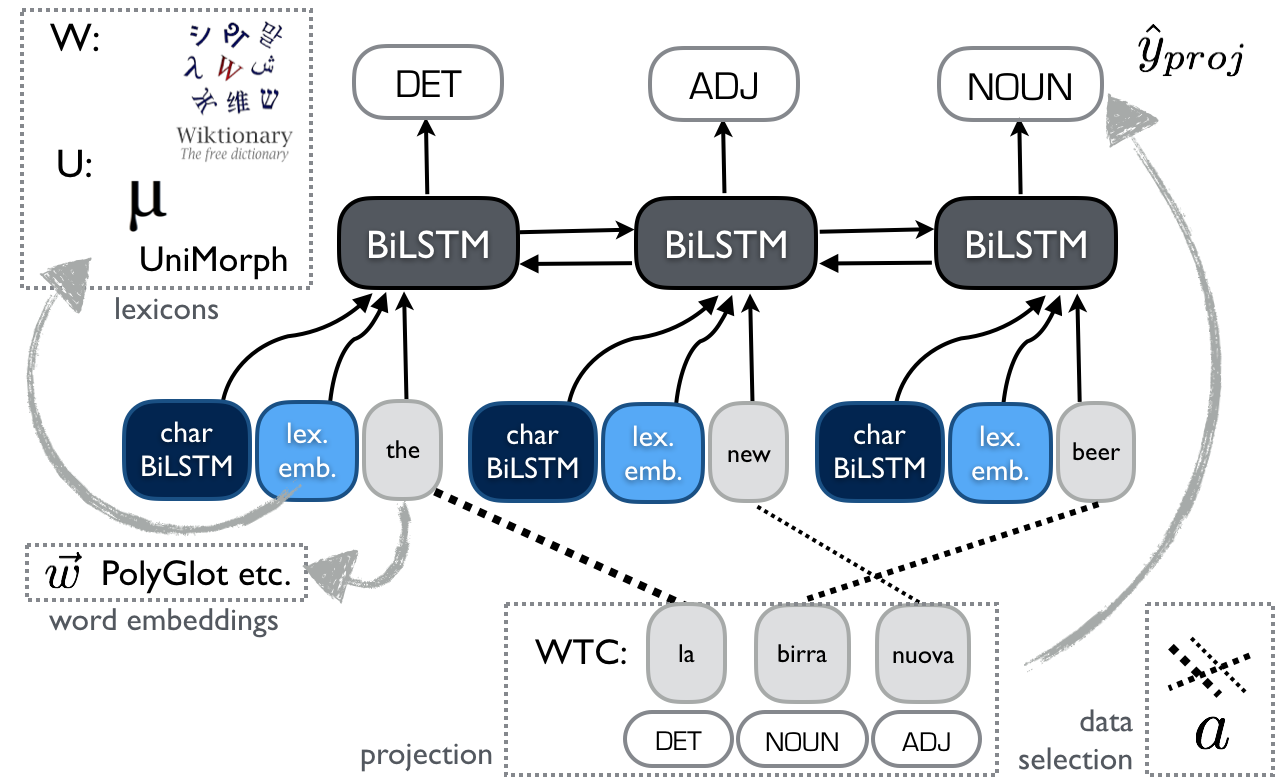}
\caption{Illustration of \textsc{DsDs} (Distant Supervision from Disparate Sources).}
\label{fig:model}
\end{figure}

\section{Method}
{\sc DsDs} is illustrated in Figure~\ref{fig:model}. The base model is a bidirectional long short-term memory network (bi-LSTM)~\cite{graves:schmidhuber:2005,Hochreiter:Schmidhuber:97,plank:ea:2016,kiperwasser2016}. 
Let $x_{1:n}$ be a given sequence of input vectors. In our base model, the input sequence consists of word embeddings $\vec{w}$ and the two output states of a character-level bi-LSTM $\vec{c}$. Given $x_{1:n}$ and a
desired index {$i$}, the function $BiRNN_\theta(x_{1:n}, i)$ (here instantiated as LSTM) reads the input sequence in forward and reverse order, respectively, and uses the concatenated ($\circ$) output states as input for tag prediction at position $i$.\footnote{CRF decoding did not consistently improve POS accuracy, as recently also independently found~\cite{yang:2018:coling}.} Our model differs from prior work on the type of input vectors $x_{1:n}$ and distant data sources, in particular, we extend the input with lexicon embeddings, all described next.

\paragraph{Annotation projection.}

Ever since the seminal work of~\newcite{yarowsky2001inducing}, projecting sequential labels from source to target languages has been one of the most prevalent approaches to cross-lingual learning. Its only requirement is that parallel texts are available between the languages, and that the source side is annotated for POS.

We apply the approach by~\newcite{agic2016multilingual}, where labels are projected from multiple sources and then decoded through weighted majority voting with word alignment probabilities and source POS tagger confidences. We exploit their wide-coverage Watchtower corpus (WTC), in contrast to the typically used Europarl data. Europarl covers 21 languages of the EU with 400k-2M sentence pairs, while WTC spans 300+ widely diverse languages with only 10-100k pairs, in effect sacrificing depth for breadth, and introducing a more radical domain shift. However, as our results show little projected data turns out to be the most beneficial, reinforcing breadth for depth.

While \citet{agic2016multilingual} selected 20k projected sentences at random to train taggers, we propose a novel alternative: selection by {\em coverage}. We rank the target sentences by percentage of words covered by word alignment from 21 sources of~\newcite{agic2016multilingual}, and select the top $k$ covered instances for training. In specific, we employ the mean coverage ranking of target sentences, whereby each target sentence is coupled with the arithmetic mean of the 21 individual word alignment coverages for each of the 21 source-language sentences. We show that this simple approach to instance selection offers substantial improvements: across all languages, we learn better taggers with significantly fewer training instances.

\paragraph{Dictionaries.}

Dictionaries are a useful source for distant supervision~\cite{li-et-al:2012,tackstrom:ea:2013}. There are several ways to exploit such information: i) as type constraints during encoding~\cite{tackstrom:ea:2013}, ii) to guide unsupervised learning~\cite{li-et-al:2012}, or iii) as additional signal at training. We focus on the latter and evaluate two ways to integrate lexical knowledge into neural models, while comparing to the former two: a) by representing lexicon properties as $n$-hot vector (e.g., if a word has two properties according to lexicon $src$, it results in a 2-hot vector, if the word is not present in $src$, a zero vector), with $m$ the number of lexicon properties; b) by \textit{embedding} the lexical features, i.e., $\vec{e}_{src}$ is a lexicon $src$ embedded into an $l$-dimensional space. We represent $\vec{e}_{src}$ as concatenation of all embedded $m$ properties of length $l$, and a zero vector otherwise. Tuning on the dev set, we found the second embedding approach to perform best, and simple concatenation outperformed mean vector representations. 

We evaluate two dictionary sources, motivated by ease of accessibility to many languages: \textsc{Wiktionary}, a word type dictionary that maps tokens to one of the 12 Universal POS tags~\cite{li-et-al:2012,petrov:ea:2012}; and \textsc{UniMorph}, a morphological dictionary that provides  inflectional paradigms across 350 languages~\citep{KIROV16.1077}. For Wiktionary, we use the freely available dictionaries from~\newcite{li-et-al:2012} and \newcite{agic:ea:2017}. The size of the dictionaries ranges from a few thousands (e.g., Hindi and Bulgarian) to 2M (Finnish UniMorph). Sizes are provided in Table~\ref{tbl:results}, first columns. UniMorph covers between 8-38 morphological properties (for English and Finnish, respectively).

\paragraph{Word embeddings.}

Embeddings are available for many languages. Pre-initialization of $\vec{w}$ offers consistent and considerable performance improvements in our distant supervision setup (Section~\ref{sec:results}). We use off-the-shelf Polyglot embeddings~\cite{polyglot}, which performed consistently better than FastText~\cite{bojanowski2016enriching}.

\section{Experiments}
\label{sec:experiments}

\paragraph{Baselines.} We compare to the following weakly-supervised POS taggers:
\begin{itemize}[noitemsep,leftmargin=*,label=--,topsep=0.25ex]
\item \textsc{\textbf{Agic:}} Multi-source annotation projection with Bible parallel data by \citet{agic:ea:2015}.
\item \textsc{\textbf{Das:}} The label propagation approach by \citet{das-petrov:2011} over Europarl data.
\item \textsc{\textbf{Garrette:}} The approach by \citet{garrette-baldridge:2013:NAACL-HLT} that works with projections, dictionaries, and unlabeled target text.
\item \textsc{\textbf{Li:}} Wiktionary supervision~\citep{li-et-al:2012}.
\end{itemize}

\paragraph{Data.} Our set of 25 languages is motivated by accessibility to embeddings and dictionaries. In all experiments we work with the 12 Universal POS tags~\cite{petrov:ea:2012}. For development, we use 21 dev sets of the Universal Dependencies 2.1 \cite{ud21}. We employ UD test sets on additional languages as well as the test sets of \citet{agic:ea:2015} to facilitate comparisons. Their test sets are a mixture of CoNLL \cite{buchholz-marsi:2006:CoNLL-X,nivre-EtAl:2007:EMNLP-CoNLL2007} and HamleDT test data \cite{zeman2014hamledt}, and are more distant from the training and development data. 

\paragraph{Model and parameters.} We extend an off-the-shelf state-of-the-art bi-LSTM tagger with lexicon information. The code is available at: \url{https://github.com/bplank/bilstm-aux}. The parameter $l$=$40$ was set on dev data across all languages. Besides using 10 epochs, word dropout rate ($p$=$.25$) and 40-dimensional lexicon embeddings, we use the parameters from~\newcite{plank:ea:2016}. 
For all experiments, we average over 3 randomly seeded runs, and provide mean accuracy. For the learning curve, we average over 5 random samples with 3 runs each.

\section{Results}\label{sec:results}

\begin{figure}
\captionsetup{farskip=0pt}
\centering
\subfloat[sentence selection]{\includegraphics[width=0.499\columnwidth]{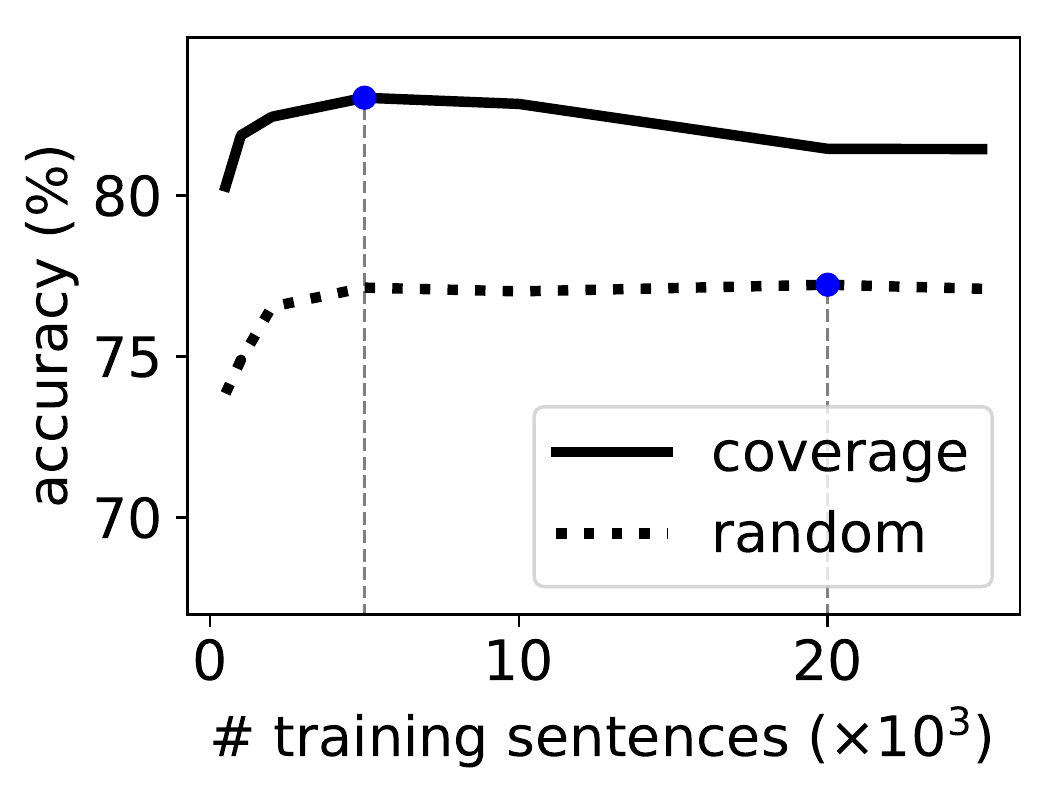}}
\subfloat[pre-trained embeddings]{\includegraphics[width=0.499\columnwidth]{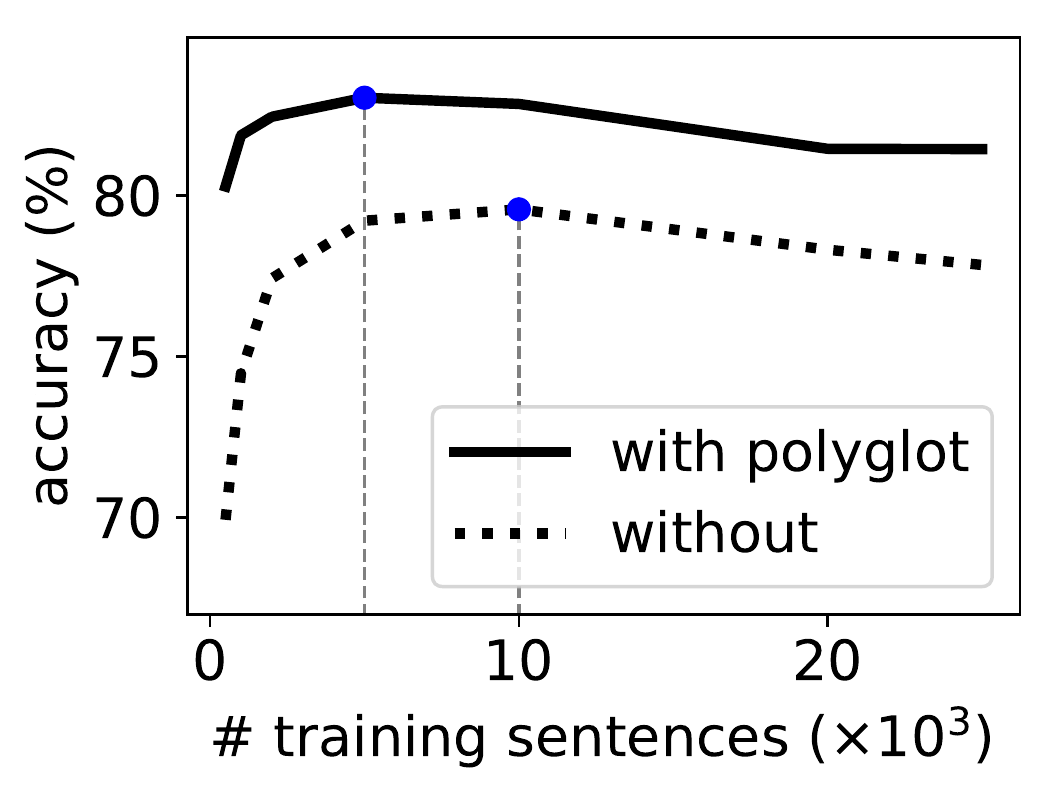}}
\caption{\label{fig:projsents}Learning curves for: a) random vs.  coverage-based sentence selection in annotation projection, both with Polyglot embeddings, and b) pre-trained embeddings on top of coverage-based selection. Means over 21 languages.}
\end{figure}

\begin{table*}[ht!]
\centering
\resizebox{\textwidth}{!}{
\begin{tabular}{l|rr|rrrrrrrrrrrr}
\toprule
& \multicolumn{2}{c|}{\textsc{Lex} ($10^3$)} & \multicolumn{5}{c}{\textsc{Dev sets (UD2.1)}}            & &  \multicolumn{5}{c}{\textsc{Test sets}} \\
\textsc{Language} & \textsc{W} & \textsc{U} & 5k & TC$_W$ & $n$-hot$_W$& $\vec{e}_W$ & \textsc{DsDs} & & \textsc{Das} & \textsc{Li} & \textsc{Garrette} & \textsc{Agic} & \textsc{DsDs} \\
\midrule
Bulgarian (bg) & 3 & 47 & 88.6 & 88.6 & 88.9 & 89.6 & \textbf{89.7} && -- & -- & 83.1 & 77.7 & \textbf{83.9}\\
Croatian (hr) & 20 & -- & 84.9 & \textbf{85.4} & 84.9 &  84.8 & $^\dagger$84.8 & &-- & -- & -- & 67.1 & $^\dagger$\textbf{78.0}\\
Czech (cs) & 14 & 72 & 86.6	& 86.6	& 86.9	& \textbf{87.6} &	87.2 && -- & -- & --   & 73.3 & \textbf{86.8} \\
Danish (da) & 22 & 24 & 89.6 & 89.0 & 89.8 & \textbf{90.2} & 90.0 && 83.2 & 83.3 & 78.8 & 79.0 & \textbf{84.5} \\
Dutch (nl) & 52 & 26 & 88.3 & 88.9 & 89.0 & 89.7 & \textbf{89.8} && 79.5 & \textbf{86.3} & -- & -- & 83.9\\
English (en) & 358 & 91 & 86.5 & \textbf{87.4}  & 86.8 & 87.3 &  87.3 & &-- & \textbf{87.1} & 80.8 & 73.0 & 85.7 \\
Finnish (fi) & 104 & 2,345 & 81.5 & 81.2 & 81.8 & 82.4 & \textbf{82.4} & & -- & -- & --& -- & --\\
French (fr) & 17 & 274 & 91.0 & 89.6 & \textbf{91.7} &  91.2 & 91.4 && -- & -- & 85.5 & 76.6 & \textbf{88.7}\\
German (de) & 62 & 71 & 85.0 & 86.4	& 85.5	& 86.0	& \textbf{86.7} && 82.8 & 85.8 & \textbf{87.1} & 80.2 & 84.1\\
Greek (el) & 21 & -- & 80.6 & \textbf{85.7} & 80.2 & 80.5 & $^\dagger$80.5 && \textbf{82.5} & 79.2 & 64.4 & 52.3 & $^\dagger$81.1\\
Hebrew (he) & 3 & 12 & 76.0 & \textbf{76.1} & 75.5 & 74.9 & 75.3 && -- & -- & -- & -- & -- \\
Hindi (hi) & 2 & 26 & 64.6 & 64.6 & 64.8 & 65.4 & \textbf{66.2} && -- & -- & -- & \textbf{67.6} &  63.1\\
Hungarian (hu) & 13 & 13 & 75.6 & 75.6 & 75.3 & 75.7 & \textbf{77.9} && -- & -- & \textbf{77.9} & 72.0 & 77.3\\
Italian (it) & 478 & 410 & 91.9 & 91.7 & 93.4 & 93.5 & \textbf{93.7} && 86.8 & 86.5 & 83.5 & 76.9 & \textbf{92.1}\\
Norwegian (no) & 47 & 18 & 90.9 & 90.9 & 90.9 & 91.0 & \textbf{91.5} && -- & -- & 84.3 & 76.7 & \textbf{86.2}\\
Persian (fa) & 4 & 26 & 42.8 & 43.0 & 43.7 & 43.5 & \textbf{43.8} && -- & -- & -- & \textbf{59.6} & 43.6 \\ 
Polish (pl) & 6 & 132 & 84.7 & 84.6 & 84.2 & 84.8 & \textbf{86.0} && -- & -- & -- & 75.1 & \textbf{84.4}\\
Portuguese & 41 & 211 & 91.4 & 91.5 & 92.3 & \textbf{92.9} & 92.2 & &87.9 & 84.5 & 87.3 & 83.8 & \textbf{89.4} \\
Romanian (ro) & 7 & 4 & 83.9 & 83.9 & 84.8 & 85.3 & \textbf{86.3} && -- & -- & --  & -- & --\\
Spanish (es) & 234 & 324 & 90.4 & 88.6 & 91.0 & 91.5 & \textbf{92.0} & & 84.2 & 86.4 & 88.7 & 81.4 & \textbf{91.7}\\
Swedish (sv) & 89 & 67 & 88.9 & 88.9 & 89.6 & \textbf{89.9} &  \textbf{89.9} & & 80.5 & \textbf{86.1} & 76.1 & 75.2 & 83.1\\
\midrule
 \textsc{Avg(21)} & & & 83.0 & 83.2 & 83.4 & 83.7 & \textbf{84.0} & \textsc{Avg(8: Das)} & 83.4 & 84.8 & 80.8 & 75.5 & \textbf{86.2} \\
 & & & & & &  &  & \textsc{Avg(8: Li$\cap$Agic)} & -- & 84.9 & 80.8 & 75.2 & \textbf{87.2}\\
\midrule
\textsc{Germanic (6)} & & & 88.2	&88.6	&88.6	&89.0	&\textbf{89.2} & \textsc{Germanic} (4: {\sc Das}) & 81.5 & \textbf{85.4} & -- & -- & 83.9\\
\textsc{Romance (5)} & & & 89.7	& 89.0	& 90.6	& 90.9	& \textbf{91.1} & \textsc{Romance} (3: {\sc Das}) & 86.3 & 85.8 & 86.5 & 80.7 & \textbf{91.1}\\
\textsc{Slavic (4)} & & & 86.2	& 86.3	& 86.2	& 86.7	& \textbf{86.9}\\
\textsc{Indo-Iranian (2)} & & & 53.7	&53.8	&54.3	&54.4	& \textbf{55.0} \\
\textsc{Uralic (2)} &  & & 78.5	& 78.4	& 78.6 &	79.0	& \textbf{80.1}\\
\bottomrule
\end{tabular}
}
\caption{Results on the development sets and comparison of our best model to prior work. \textsc{Lex}: Size (word types) of dictionaries (W: Wiktionary, U: UniMorph). TC$_W$: type-constraints using Wiktionary; $\vec{e}_W$ (embedded Wiktionary tags), \textsc{DsDs}: our model with $\vec{e}_{W \cup U}$. Results indicated by $^\dagger$ use W only. Best result in boldface; in case of equal means, the one with lower std is boldfaced. Averages over language families (with two or more languages in the sample, number of languages in parenthesis). } 
\label{tbl:results}
\end{table*}

Table~\ref{tbl:results} shows the tagging accuracy for individual languages, while the means over all languages are given in Figure~\ref{fig:projsents}. There are several take-aways.

\paragraph{Data selection.} The first take-away is that coverage-based instance selection yields substantially better training data. Most prior work on annotation projection resorts to arbitrary selection; informed selection clearly helps in this noisy data setup, as shown in Figure~\ref{fig:projsents} (a). Training on 5k instances results in a sweet spot; more data (10k) starts to decrease performance, at a cost of  runtime. Training on all WTC data (around 120k) is worse for most languages. From now on we consider the 5k model trained with Polyglot as our baseline (Table~\ref{tbl:results}, column ``5k''), obtaining a mean accuracy of 83.0 over 21 languages.

\paragraph{Embeddings initialization.} Polyglot initialization offers a large boost; on average $+$3.8\% absolute improvement in accuracy for our 5k training scheme, as shown in Figure~\ref{fig:projsents} (b). 
The big gap in low-resource setups further shows their effectiveness, with up to 10\% absolute increase in accuracy when training on only 500 instances.

\paragraph{Lexical information.} The main take-away is that lexical information helps neural tagging, and \textit{embedding} it proves the most helpful. Embedding Wiktionary tags reaches 83.7 accuracy on average, versus 83.4 for $n$-hot encoding, and 83.2 for type constraints. Only on 4 out of 21 languages are type constraints  better. This is the case for only one language for $n$-hot encoding (French). The best approach is to embed both Wiktionary and Unimorph, boosting performance further to 84.0, and resulting in our final model. {It helps the most on morphological rich languages such as Uralic.

On the test sets (Table~\ref{sec:results}, right) {\sc DsDs} reaches 87.2 over 8 test languages intersecting~\newcite{li-et-al:2012} and~\newcite{agic2016multilingual}. It reaches 86.2 over the more commonly used 8 languages of~\citet{das-petrov:2011}, compared to their 83.4. This shows that our novel ``soft'' inclusion of noisy dictionaries is superior to a hard decoding restriction, and including lexicons in neural taggers helps. We did not assume any gold data to further enrich the lexicons, nor fix possible tagset divergences.

\section{Discussion}
\paragraph{Analysis.} The inclusion of lexicons results in higher coverage and is part of the explanation for the improvement of {\sc DsDs}; see correlation in Figure~\ref{fig:details} (a). What is more interesting is that our model benefits from the lexicon beyond its content: OOV accuracy for words \textit{not} present in the lexicon overall improves, besides the expected improvement on known OOV, see Figure~\ref{fig:details} (b).

\begin{table*}
\centering
\resizebox{\textwidth}{!}{
\begin{tabular}{llll|rr|cccccr}
\toprule
            & & &   & \multicolumn{2}{c|}{\textsc{Lex} ($10^3$)} & \multicolumn{6}{c}{{\sc Test sets}}\\
\textsc{Language} & \textsc{Test} & \textsc{Proj} & \textsc{Emb} & W & U & TnT & 5k & TC$_W$ & $n$-hot$_W$& $\vec{e}_W$ & \textsc{DsDs}  \\
\midrule
Basque (eu) & UD & Bible & eu & 1 & -- &  57.5 & 61.8 & 61.8 & 61.4 & \textbf{62.7} &$\dagger$\textbf{62.7}\\
Basque (eu) & CoNLL & Bible & eu & 1 & -- & 57.0 & 60.3 & 60.3 & 60.3 & \textbf{61.3} & $\dagger$\textbf{61.3} \\

Estonian (et) & UD & WTC & et & -- & 10 &  79.5 &80.6 & -- & -- & -- & \textbf{81.5}\\
Serbian (sr) & UD & WTC (hr) & hr & (hr) 20 & -- & 84.0& 84.7 &	\textbf{85.5} &	85.1 &	85.2 & $\dagger$85.2\\ 
Serbian (sr) & UD & Bible (sr) & hr  & (hr) 20 & --& 77.1 & 78.9 & 79.4 & 80.5 & \textbf{80.7} & $\dagger$\textbf{80.7}\\
Tamil (ta) & UD & WTC & ta &  -- & -- & 58.2 & \textbf{61.2} & --  & -- & -- & -- \\
\bottomrule
\end{tabular}
}
\caption{Results for languages with missing data sources: WTC projections, Wiktionary (W), or UniMorph (U). Test sets ({\sc Test}), projection sources ({\sc Proj}), and embeddings languages ({\sc Emb}) are indicated. Comparison to TnT~\cite{Brants2000} trained on {\sc Proj}. Results indicated by $^\dagger$ use W only.}
\label{tbl:closelowlang}
\end{table*}

\paragraph{More languages.} All data sources employed in our experiment are very high-coverage. However, for true low-resource languages, we cannot safely assume the availability of all disparate information sources. Table~\ref{tbl:closelowlang} presents results for four additional languages where some supervision sources are missing. We observe that adding lexicon information always helps, even in cases where only 1k entries are available, and embedding it is usually the most beneficial way. For closely-related languages such as Serbian and Croatian, using resources for one aids tagging the other, and modern resources are a better fit. For example, using the Croatian WTC projections to train a model for Serbian is preferable over in-language Serbian Bible data where the OOV rate is much higher.

\begin{figure}[t]
\captionsetup{farskip=0pt}
\centering
\subfloat[coverage vs.\ accuracy]
{\includegraphics[width=0.499\columnwidth,height=7.35em]{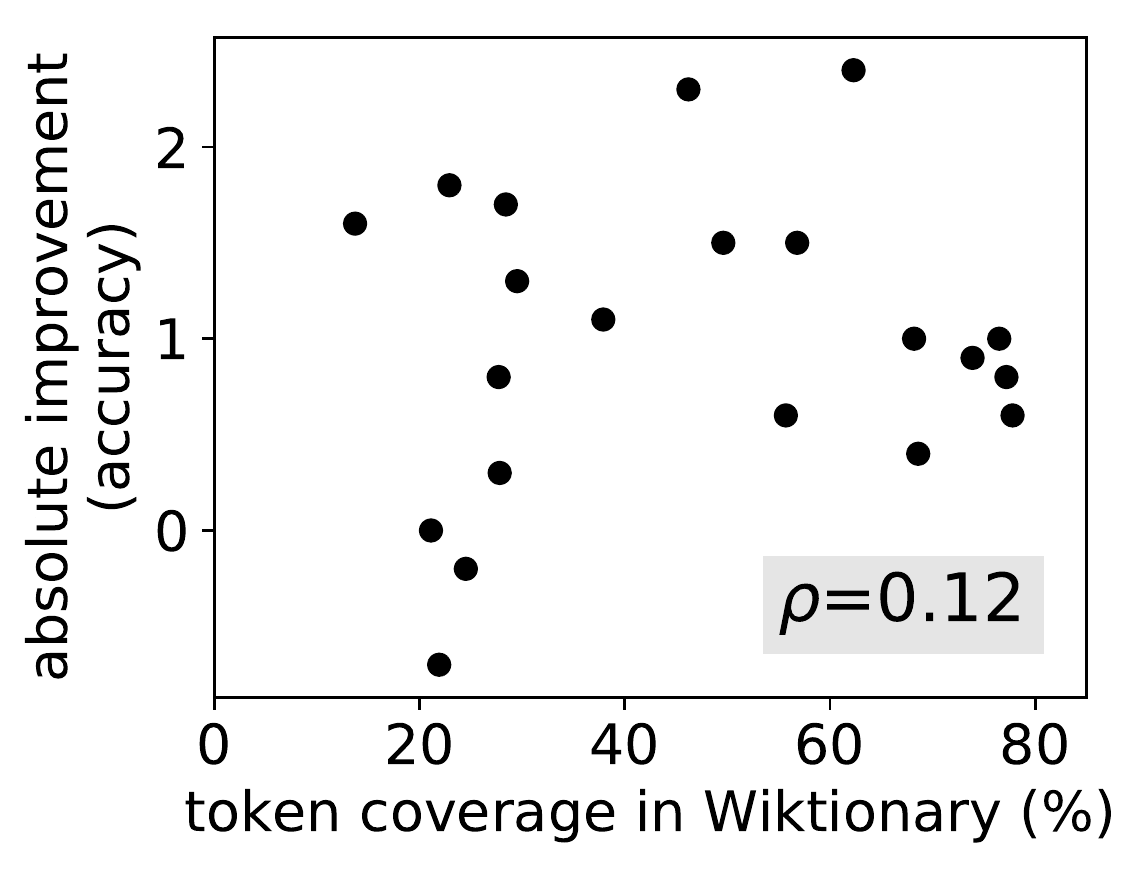}}
\subfloat[OOV accuracy]{\includegraphics[width=0.499\columnwidth,height=7.35em]{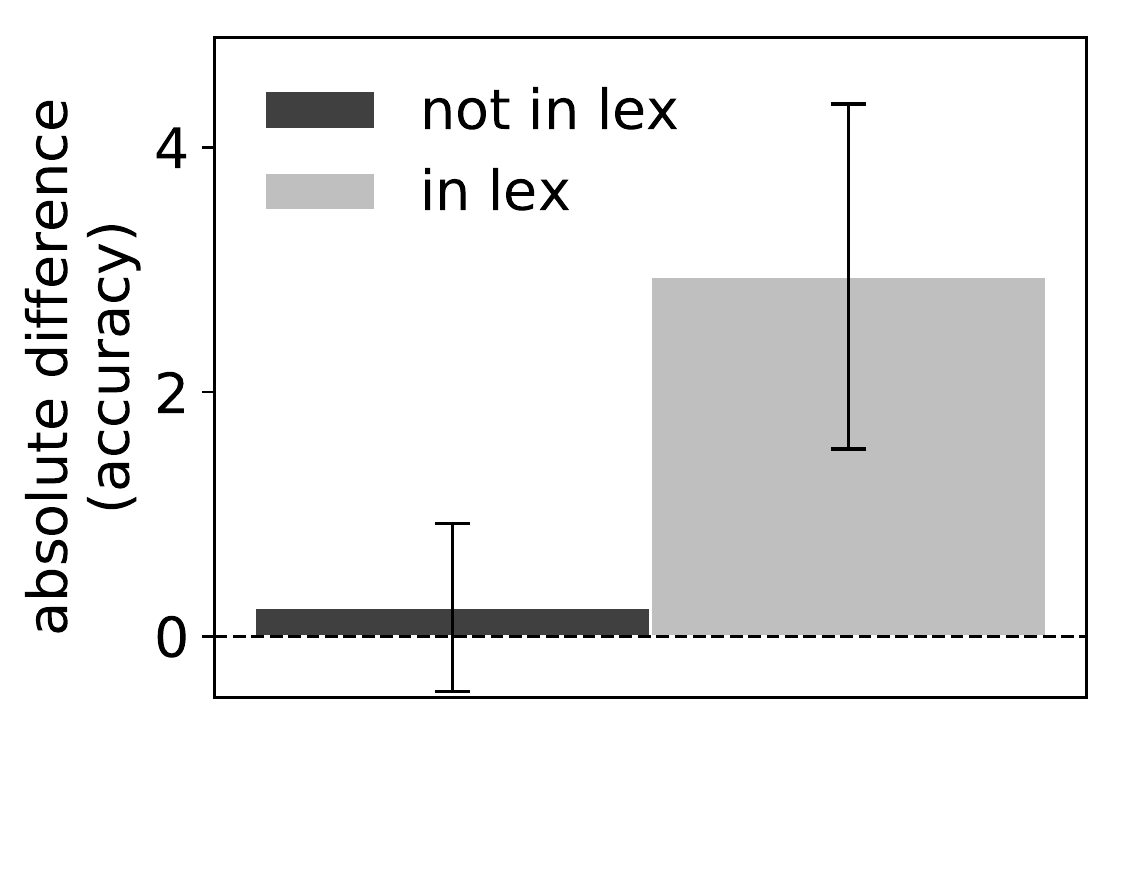}}
\caption{\label{fig:details} Analysis of {\sc DsDs} accuracy improvements over the baseline on all development languages with respect to a) token coverage by the lexicon, including Pearson's $\rho$; b) OOV accuracy for tokens in/not in the lexicon, with 95\% confidence intervals of the mean. Here, a token is covered if we can find it in at least one lexicon.}
\end{figure}
\begin{figure}
\end{figure}

\paragraph{How much gold data?} We assume not having access to \textit{any} gold annotated data. It is thus interesting to ask how much gold data is needed to reach our performance. This is a tricky question, as training within the same corpus naturally favors the same corpus data. We test both in-corpus (UD) and out-of-corpus data (our test sets) and notice an important gap: while in-corpus only 50 sentences are sufficient, outside the corpus one would need over 200 sentences. This experiment was done for a subset of 18 languages with both in- and out-of-corpus test data.

\paragraph{Further comparison.} In Table~\ref{tbl:results} we directly report the accuracies from the original contributions by {\sc Das, Li, Garrette}, and {\sc Agic} over the same test data. We additionally attempted to reach the scores of {\sc Li} by running their tagger over the Table~\ref{tbl:results} data setup. The results are depicted in Figure~\ref{fig:iters} as mean accuracies over EM iterations until convergence. We show: i) {\sc Li} peaks at 10 iterations for their test languages, and at 35 iterations for all the rest. This is in slight contrast to 50 iterations that~\citet{li-et-al:2012} recommend, although selecting 50 does not dramatically hurt the scores; ii) Our replication falls $\sim$5 points short of their 84.9 accuracy. There is a large 33-point accuracy gap between the scores of~\citet{li-et-al:2012}, where the dictionaries are large, and the other languages in Figure~\ref{fig:iters}, with smaller dictionaries.

Compared to {\sc Das}, our tagger clearly benefits from pre-trained word embeddings, while theirs relies on label propagation through Europarl, a much cleaner corpus that lacks the coverage of the noisier WTC. Similar applies to~\citet{tackstrom:ea:2013}, as they use 1-5M near-perfect parallel sentences. Even if we use much smaller and noisier data sources, {\sc DsDs} is almost on par: 86.2 vs. 87.3 for the 8 languages from~\citet{das-petrov:2011}, and we even outperform theirs on four languages: Czech, French, Italian, and Spanish.

\begin{figure}[t]
\centering
\includegraphics[width=\columnwidth]{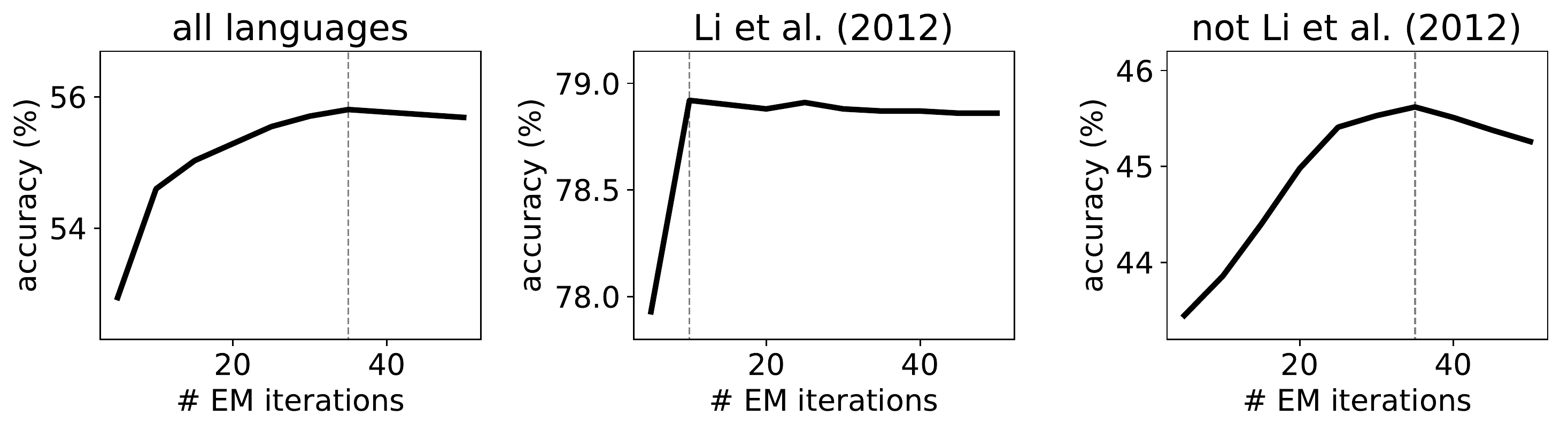}
\caption{\label{fig:iters} The performance of {\sc Li} with our dictionary data over EM iterations, separate for the languages from~\citet{li-et-al:2012} and all the remaining languages in Table~\ref{tbl:results}.}
\end{figure}
\begin{figure}
\end{figure}

\section{Related Work}
Most successful work on low-resource POS tagging is based on projection~\cite{yarowsky2001inducing}, tag dictionaries~\citep{li-et-al:2012}, annotation of seed training data~\cite{garrette-baldridge:2013:NAACL-HLT} or even more recently some combination of these, e.g., via multi-task learning~\cite{fang2016learning,kann:ea:2018}. Our paper contributes to this literature by leveraging a range of prior directions in a unified, neural test bed. 

Most prior work on neural sequence prediction follows the commonly perceived wisdom that hand-crafted features are unnecessary for deep learning methods. They rely on end-to-end training without resorting to additional linguistic resources. Our study shows that this is not the case. Only few prior studies investigate such sources, e.g., for MT~\cite{sennrich-haddow:2016:WMT,chen:ea:2017,Li:ea:2017,passban:ea:2018} and~\newcite{sagot-martinezalonso:2017:IWPT}  for POS tagging use lexicons, but only as $n$-hot features and without examining the cross-lingual aspect.

\section{Conclusions}

We show that our approach of distant supervision from disparate sources (\textsc{DsDs}) is simple yet surprisingly effective for low-resource POS tagging. Only 5k instances of projected data paired with off-the-shelf embeddings and lexical information integrated into a neural tagger are sufficient to reach a new state of the art, and both data selection and embeddings are essential components to boost neural tagging performance. 

\bibliography{biblio}
\bibliographystyle{acl_natbib_nourl}

\end{document}